\documentclass[11pt,a4paper]{article}
\usepackage{authblk}
\usepackage[hyperref]{emnlp2018}
\usepackage{breakurl}
\usepackage{times}
\usepackage{latexsym}

\usepackage{url}
\usepackage{booktabs}
\usepackage{multirow}
\usepackage{amsmath}
\usepackage{amssymb}
\newcommand{\R}{\mathbb{R}}
\usepackage[T1]{fontenc}
\usepackage{color}
\usepackage{bm}
\usepackage{graphicx}
\usepackage{subcaption}
\usepackage{seqsplit}
\usepackage{pgfplots}
\usepackage[printwatermark]{xwatermark}
\pgfplotsset{compat=newest}
\usepgfplotslibrary{colorbrewer}
\pgfplotsset{colormap/RdBu-9}
\usepackage{array}
\newcolumntype{C}[1]{>{\centering\let\newline\\\arraybackslash\hspace{0pt}}m{#1}}
\pgfmathdeclarefunction{gauss}{2}{%
  \pgfmathparse{ 1/(#2*sqrt(2*pi))*exp(-((x-#1)^2)/(2*#2^2)) }%
}
\def\addlegendimage{\csname pgfplots@addlegendimage\endcsname}
\usepackage{soul}

\usepackage{todonotes}

\aclfinalcopy 

\title{Explaining Character-Aware Neural Networks for Word-Level Prediction:\\
Do They Discover Linguistic Rules?}

\author{\textbf{Fr\'{e}deric Godin, Kris Demuynck, Joni Dambre, Wesley De Neve and Thomas Demeester }}
\affil{IDLab, Ghent University - imec, Ghent, Belgium\\
\tt{firstname.lastname@ugent.be}}

\date{}

\begin{document}
\maketitle
\begin{abstract}
Character-level features are currently used in different neural network-based natural language processing algorithms. However, little is known about the character-level patterns those models learn. Moreover, models are often compared only quantitatively while a qualitative analysis is missing. In this paper, we investigate which character-level patterns neural networks learn and if those patterns coincide with manually-defined word segmentations and annotations. To that end, we extend the contextual decomposition \cite{james2018beyond} technique to convolutional neural networks which allows us to compare convolutional neural networks and bidirectional long short-term memory networks. We evaluate and compare these models for the task of morphological tagging on three morphologically different languages and show that these models implicitly discover understandable linguistic rules.

\end{abstract}

\section{Introduction}

Character-level features are an essential part of many Natural Language Processing (NLP) tasks. These features are for instance used for language modeling \cite{Kim_char}, part-of-speech tagging \cite{plank} and machine translation \cite{luong2016acl_hybrid}. They are especially useful in the context of part-of-speech and morphological tagging, where for example the suffix \emph{-s} can easily differentiate plural words from singular words in English or Spanish.

The use of character-level features is not new. Rule-based taggers were amongst the earliest systems that used character-level features/rules for grammatical tagging \cite{Klein:1963:CAG:321172.321180}. Other approaches rely on fixed lists of affixes \cite{Ratnaparkhi96amaximum,Toutanova:2003:FPT:1073445.1073478}. Next, these features are used by a tagging model, such as a rule-based model or statistical model. Rule-based taggers are transparent models that allow us to easily trace back why the tagger made a certain decision (e.g., \citet{Brill:1994:RRP:1075812.1075869}). Similarly, statistical models are merely a weighted sum of features.

For example, \citet{Brill:1994:RRP:1075812.1075869}'s transformation-based error-driven tagger uses a set of templates to derive rules by fixing errors. The following rule template: 

\vspace{3 mm}
\emph{"Change the most-likely tag \textbf{X} to \textbf{Y} if the last (1,2,3,4) characters of the word are \textbf{x}"},
\vspace{3 mm}

resulted in the rule:

\vspace{3 mm}
\emph{"Change the tag \textbf{common noun} to \textbf{plural common noun} if the word has suffix \textbf{-s}"}.
\vspace{3 mm}

Subsequently, whenever the tagger makes a tagging mistake, it is easy to trace back why this happened. Following the above rule, the word \emph{mistress} will mistakingly be tagged as a plural common noun while it actually is a common noun\footnote{In \citet{Brill:1994:RRP:1075812.1075869}, an additional rule encodes an exception to this rule to correctly tag the word \emph{mistress}.}. 

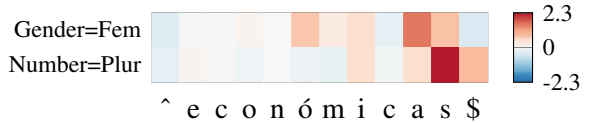
\begin{figure}
	\centering
    \begin{tikzpicture}
\begin{axis}[ymin=-0.5,ymax=1.5,xmin=-0.5,xmax=11.5,point meta min = -1,xtick={0,1,2,3,4,5,6,7,8,9,10,11},point meta max = 1,yticklabels={\small Number=Plur,\small Gender=Fem},ytick={0,1},xticklabels={\^{},e,c,o,n,\'{o},m,i,c,a,s,\$},height=2.5cm,width=6cm, colorbar, colorbar style={
		width=0.25cm,
        ticklabel style = {font=\small},
        ytick={-1,0,1},
        yticklabels={-2.3,0,2.3} },   
colormap={reverse rgbu}{
        indices of colormap={
            \pgfplotscolormaplastindexof{RdBu-9},...,0 of {RdBu-9}}
    },typeset ticklabels with strut]
\addplot [matrix plot*,point meta=explicit] file [meta=index 2] {figure_intro.dat};
\end{axis}
\end{tikzpicture}
    \caption{Individual character contributions of the Spanish adjective \emph{econ\'{o}micas}. The character \emph{a} has the highest positive (red) contribution for predicting the label \emph{Gender=Fem}, and the character \emph{s} for predicting the label \emph{Number=Plur}. This coincides with our linguistic knowledge of Spanish. }
    \label{fig:figure_intro}
\end{figure}

This is in stark contrast with the most recent generation of part-of-speech and morphological taggers which mainly rely on neural networks. Words are split into individual characters and are in general either aggregated using a Bidirectional Long Short-Term Memory network (BiLSTM) \cite{plank} or Convolutional Neural Network (CNN) \cite{DBLP:conf/icml/SantosZ14}. However, it is currently unknown which character-level patterns these neural network models learn and whether these patterns coincide with our linguistic knowledge. Moreover, different neural network architectures are currently only compared quantitatively and lack a qualitative analysis.

In this paper, we investigate which character patterns neural networks learn and 
to what extent those patterns comprise any known linguistic rules. 
We do this for three morphologically different languages: Finnish, Spanish and Swedish.
A Spanish example is shown in Figure~\ref{fig:figure_intro}. By visualizing the contributions of each character, we observe that the model indeed uses the suffix \emph{-s} to correctly predict that the word is plural.

Our main contributions are as follows:
\begin{itemize}
\item We show how word-level tagging decisions can be
traced back
to specific sets of characters and interactions between them. 
\item We extend the contextual decomposition method \cite{james2018beyond} to CNNs.
\item We quantitatively compare CNN and BiLSTM models in the context of morphological tagging by performing an evaluation on three manually segmented and morphologically annotated corpora. 

\item We found out that the studied neural models are able to implicitly discover character patterns that coincide with the same rules linguists use to indicate the morphological function of subword segments.

\end{itemize}

Our implementation is available online\footnote{\url{https://github.com/FredericGodin/ContextualDecomposition-NLP}}.

\section{Related Work}
Neural network-based taggers currently outperform statistical taggers in morphological tagging \cite{heigold} and part-of-speech tagging \cite{plank} for a wide variety of languages. Character-level features form a crucial part of many of these systems. Generally, two neural network architectures are considered for aggregating the individual characters: a BiLSTM \cite{ling_char,plank} or a CNN \cite{DBLP:conf/icml/SantosZ14,C16-1333,heigold}. These architectures outperform similar models that use manually defined features \cite{ling_char,DBLP:conf/icml/SantosZ14}. However, it is still unclear which useful character-level features they have learned. Architectures are compared quantitatively but lack insight into learned patterns. Moreover, \citet{vania} showed in the context of language modeling that training a BiLSTM on ground truth morphological features still yields better results than eight other character-based neural network architectures. Hence, this raises the question which patterns neural networks learn and whether these patterns coincide with manually-defined linguistic rules. 

While a number of interpretation techniques have been proposed for images \cite{DBLP:journals/corr/SpringenbergDBR14,Selvaraju_2017_ICCV,pmlr-v70-shrikumar17a}, these are generally not applicable in the context of NLP where LSTMs are mainly used. Moreover, gradient-based techniques are not trustworthy when strongly saturating activation functions  such as \emph{tanh} and \emph{sigmoid} are used (e.g., \citet{N16-1082}). Hence, current interpretations in NLP are limited to visualizing the magnitude of the LSTM hidden states of each word \cite{linzen,DBLP:journals/corr/RadfordJS17,DBLP:journals/tvcg/StrobeltGPR18}, removing words \cite{DBLP:journals/corr/LiMJ16a,kadar} or changing words \cite{linzen} and measuring the impact,  or training surrogate tasks \cite{DBLP:journals/corr/AdiKBLG16,P17-1057chrupla,P17-1080Belinkov}. These techniques only provide limited local interpretations and do not model fine-grained interactions of groups of inputs or intermediate representations. In contrast, \citet{james2018beyond} recently introduced an LSTM interpretation technique called Contextual Decomposition (CD), providing a solution to the aforementioned issues. We will build upon this interpretation technique and introduce an extension for CNNs, making it possible to compare different neural network architectures within a single interpretation framework.

\section{Method}
For visualizing the contributions of character sets, we use the recently introduced Contextual Decomposition (CD) framework, as originally developed for LSTMs \cite{james2018beyond}, and extend it to CNNs.
First, we introduce the concept of CD, followed by the extension for CNNs. For details on CD for LSTMs, we refer the reader to the aforementioned paper. Finally, we explain how the CD of the final classification layer is done.
\subsection{Contextual decomposition}
The idea behind CD is that, in the context of character-level decomposition, we can decompose the output value of the network for a certain class into two distinct groups of contributions: (1) contributions originating from a specific character or set of characters within a word and (2) contributions originating from all the other characters within the same word.

More generally, we can decompose every output value $z$ of every neural network component into a relevant contribution $\beta$ and an irrelevant contribution $\gamma$:
\begin{equation}
z = \beta + \gamma
\end{equation}

\subsection{Decomposing CNN layers}
A CNN typically consist of three components: the convolution itself, an activation function and an optional max-pooling operation. We will discuss each component in the next paragraphs.

\paragraph{Decomposing the convolution}
Given a sequence of character embeddings $\bm{x}_1,...,\bm{x}_T\in\R^{d_1}$ of length $T$, we can calculate the convolution of size $n$ of a single filter over the sequence $\bm{x}_{1:T}$ by applying the following equation to each $n$-length subsequence $\{\bm{x}_{t+i},i=0,..,n-1\}$, denoted as $\bm{x}_{t:t+n-1}$:
\begin{equation}
z_t = \sum_{i=0}^{n-1} W_i\cdot \bm{x}_{t+i} + b,
\end{equation}
with  $z_t \in\R$ and where $W \in \R^{d_1 \times n }$ and $b \in\R$ are the weight matrix and bias of the convolutional filter. $W_i$ denotes the $i$-th column of the weight matrix $W$.

When we want to calculate the contribution of a subset of characters, where $\bm{S}$ is the set of corresponding character position indexes and $\bm{S} \subseteq \{1,...,T\}$, we should decompose the output of the filter $z_t$ into three parts: 
\begin{equation}
z_t = \beta_t + \gamma_t + b.
\end{equation}
That is, the relevant contribution $\beta_t$ originating from the selected subset of characters with indexes $\bm{S}$, the irrelevant contribution $\gamma_t$ originating from the remaining characters in the sequence, and a bias which is deemed neutral \cite{james2018beyond}.

This can be achieved by decomposing the convolution itself as follows:
\begin{equation}
\beta_t = \sum_{i=0}^{n-1} W_i\cdot \bm{x}_{t+i} \quad \quad (t+i) \in \bm{S}, \\
\end{equation}
\begin{equation}
\gamma_t = \sum_{i=0}^{n-1} W_i\cdot \bm{x}_{t+i} \quad \quad (t+i) \notin \bm{S}, \\
\end{equation}

\paragraph{Linearizing the activation function}
After applying a linear transformation to the input, a non-linearity is typically applied. In CNNs, the ReLU activation function is often used. 

In \citet{james2018beyond}, a linearization method for the non-linear activation function $f$ is proposed, based on the differences of partial sums of all $N$ components $y_i$ involved in the pre-activation sum $z_t$. 
In other words, we want to split $f_{ReLU}(z_t) = f_{ReLU}(\sum_{i=1}^N y_i)$ into a sum of individual linearized contributions $L_{f_{ReLU}}(y_i)$, namely $f_{ReLU}(\sum_{i=1}^N y_i) = \sum_{i=1}^N L_{f_{ReLU}}(y_i)$. To that end, we compute $L_{f_{ReLU}}(y_k)$, the linearized contribution of $y_k$ as the average difference of partial sums over all possible permutations $\pi_1,...,\pi_{M_N}$ of all $N$ components $y_i$ involved:

\begin{multline}
L_f(y_k) = \\ \frac{1}{M_N}\sum_{i=1}^{M_N} [ f(\sum_{l=1}^{\pi_i^{-1}(k)} y_{\pi_i(l)}) - f(\sum_{l=1}^{\pi_i^{-1}(k)-1} y_{\pi_i(l)})    ]
\end{multline}

Consequently, we can decompose the output $c_t $ after the activation function as follows:
\begin{align}
c_t =& f_{ReLU}(z_t)\\
 =& f_{ReLU}(\beta_{z,t} + \gamma_{z,t} + b)\\
 =& L_{ReLU}(\beta_{z,t}) \nonumber\\
 &+ [L_{ReLU}(\gamma_{z,t}) +L_{ReLU}(b)] \\
 =&  \beta_{c,t} + \gamma_{c,t}\label{eq:decomp}
\end{align}
Following \citet{james2018beyond}, $\beta_{c,t}$ contains the contributions that can be directly attributed to the specific set of input indexes $\bm{S}$. Hence, the bias $b$ is part of $\gamma_{c,t}$. Note that, while the decomposition in Eq.\ (\ref{eq:decomp})
is exact in terms of the total sum, the individual attribution to relevant ($\beta_{c,t}$) and irrelevant ($\gamma_{c,t}$) is an approximation, due to the linearization. 

\paragraph{Max-pooling over time}
When applying a fixed-size convolution over a variable-length sequence, the output is again of variable size. Hence, a max-pooling operation is executed over the time dimension, resulting in a fixed-size representation that is independent of the sequence length:
\begin{equation}
c = \max_t(c_t).
\end{equation}
Instead of applying a max operation over the $\beta_{c,t}$ and $\gamma_{c,t}$ contributions separately, we first determine the position $t$ of the highest $c_t$ value and propagate the corresponding $\beta_{c,t}$ and $\gamma_{c,t}$ values.

\subsection{Calculating the final contribution scores}
The final layer is a classification layer, which is the same for a CNN- or LSTM-based architecture.
The probability $p_j$ of predicting class $j$ is defined as follows:
\begin{equation}\label{eq:softmax}
p_j= \frac{e^{W_j\cdot\bm{x}+b_j}}{\sum_{i=1}^C e^{W_i\cdot\bm{x}+b_i}},
\end{equation}
in which $W \in \R^{d_2 \times C }$ is a weight matrix and $W_i$ the i-th column, $\bm{x} \in \R^{d_2}$ the input, $\bm{b} \in\R^{d_2}$ the bias vector and $b_i$ the i-th element, $d_2$ the input vector size and $C$ the total number of classes. 

The input $\bm{x}$ is either the output $\bm{c}$ of a CNN or $\bm{h}$ of a LSTM. Consequently, we can decompose $\bm{x}$ into $\bm{\beta}$ and $\bm{\gamma}$ contributions. In practice, we only consider the preactivation and decompose it as follows: 
\begin{equation}
W_j\cdot\bm{x}+b_j = W_j\cdot\bm{\beta}+W_j\cdot\bm{\gamma}+b_j.
\end{equation}
Finally, the contribution of a set of characters with indexes $\bm{S}$ to the final score of class $j$ is equal to $W_j\cdot\bm{\beta}$. The latter score is used throughout the paper for visualizing contributions of sets of characters.

\section{Experimental Setup}
We execute experiments on morphological tagging in three different languages: Finnish, Spanish and Swedish. We describe the dataset in Section~\ref{sec:dataset}, whereas model and training details can be found in Section~\ref{sec:model_details}.

\subsection{Dataset} \label{sec:dataset}
For our experiments, we use the Universal Dependencies 1.4 (UD) dataset \cite{NIVRE16.348}, which contains morphological features for a large number of sentences. 
Additionally, we acquired manually-annotated character-level morphological segmentations and labels for a subset of the test set for three morphological different languages: Finnish, Spanish and Swedish. \footnote{Available online: \seqsplit{http://github.com/mpsilfve/ud-segmenter/commit/5959214d494cbc13e53e1b26650813ff950d2ee3}}

For each language, \citet{silfverberg} selected the first non-unique 300 words from the UD test set and manually segmented each word according to the associated lemma and morphological features in the dataset. Whenever possible, they assigned each feature to a specific subset of characters. For example, the Spanish word "econ\'{o}micas" is segmented as follows:

\begin{itemize}
\itemsep0em 
  \item econ\'{o}mic : lemma=econ\'{o}mico
  \item a : gender=feminine
  \item s : number=plural
\end{itemize}
For our experiments, we are only interested in word/feature pairs for which a feature can be assigned to a specific subset of characters. Hence, we filter the test set on those specific word/feature pairs. In the above example, we have two word/feature pairs. This resulted in 278, 340 and 137 word/feature pairs for Finnish, Spanish and Swedish, respectively. Using the same procedure, we selected relevant feature classes, resulting in 12, 6 and 9 feature classes for Finnish, Spanish and Swedish, respectively. For each class, when a feature was not available, we introduced an additional \emph{Not Applicable} (NA) label. A complete overview of the feature classes can be found in Appendix~\ref{app:morph_classes_used}.

We always train and validate on the full UD dataset for which we have filtered out all duplicate words. After that, we perform our analysis on either the UD test set or the annotated subset of manually segmented and annotated words. An overview can be found in Table~\ref{tab:stats}.

\begin{table}[t]
\centering
\caption{Overview of the training, validation and test set used.}
\begin{tabular}{ C{2.2cm} c c c}
\toprule
 & \textbf{Finnish} & \textbf{Spanish} & \textbf{Swedish} \\ 
 \midrule
\textbf{Train words} &53547&62556&16295\\
\textbf{Valid words} &2317&4984&1731\\
\textbf{Test words} &2246&956&3538\\
\midrule
\textbf{Annotated Test pairs} &278&340&137\\
\bottomrule
\end{tabular}
\label{tab:stats}
\end{table}

\subsection{Model} \label{sec:model_details}
We experiment with both a CNN and BiLSTM architecture for character-level modeling of words. 

At the input, we split every word into characters and add a start-of-word (\^{}) and an end-of-word (\$) character.
With every character, we associate a character embedding of size 50.

Our CNN architecture is inspired by \citet{Kim_char} and consists of a set of filters of varying width, followed by a ReLU activation function and a max-over-time pooling operation. We adopt their small-CNN parameter choices and have 25, 50, 75, 100, 125 and 150 convolutional filters of size 1, 2, 3, 4, 5 and 6, respectively. We do not add an additional highway layer.

For the character-level BiLSTM architecture, we follow the variant used in \citet{plank}. That is, we simply run a BiLSTM over all the characters and concatenate the final forward and backward hidden state. 
To obtain a similar number of parameters as the CNN model, we set the hidden state size to 100 units for each LSTM. 

Finally, the word-level representation generated by either the CNN or BiLSTM architecture is classified by a multinomial logistic regression layer. Each morphological class type has a different layer. We do not take into account context to rule out any influence originating from somewhere other than the characters of the word itself.

\paragraph{Training details} For morphological tagging, we train a single model for all classes at once. We minimize the joint loss by summing the cross-entropy losses of each class. We orthogonally initialize all weight matrices, except for the embeddings, which are uniformly initialized ([-0.01;0.01]). All models are trained using Adam \cite{journals/corr/KingmaB14} with minibatches of size 20 and learning rate 0.001. No specific regularization is used. We select our final model based on early stopping on the validation set.

\section{Experiments}
First, we verify that the CD algorithm works correctly by executing a controlled experiment with a synthetic token. Next, we quantitatively and qualitatively evaluate on the full test set.

\subsection{Validation of contextual decomposition for convolutional neural networks}

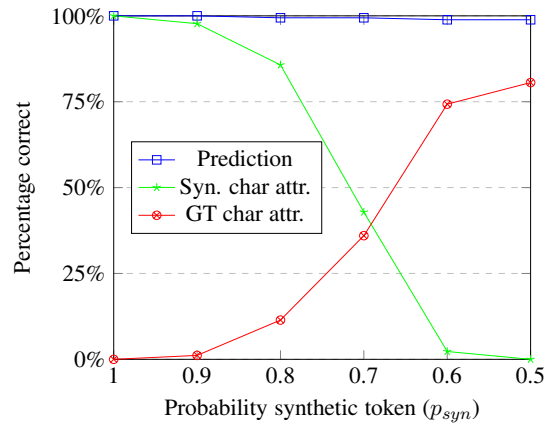
\begin{figure}
	\centering
    \begin{tikzpicture}[scale=0.8]
\begin{axis}[
    xlabel={Probability synthetic token ($p_{syn}$)},
    ylabel={Percentage correct},
    xmin=0, xmax=5,
    ymin=0, ymax=175,
    xtick={0,1,2,3,4,5},
    xticklabels={1,0.9,0.8,0.7,0.6,0.5},
    ytick={0,43.75,87.5,131.25,175},
    yticklabels={0\%,25\%,50\%,75\%,100\%},
	legend style={at={(0.5,0.5)},anchor=east},
    ymajorgrids=true,
    grid style=dashed
] 
\addplot[
    color=blue,
    mark=square,
    ]
    coordinates {
    (0,175)(1,175)(2,174)(3,174)(4,173)(5,173)
    };
    \addlegendentry{Prediction}
\addplot[
    color=green,
    mark=star,
    ]
    coordinates {
    (0,175)(1,171)(2,150)(3,75)(4,4)(5,0)
    };
    \addlegendentry{Syn. char attr.}
\addplot[
    color=red,
    mark=otimes,
    ]
    coordinates {
    (0,0)(1,2)(2,20)(3,63)(4,130)(5,141)
    };
    \addlegendentry{GT char attr.}
\end{axis}
\end{tikzpicture}
    \caption{Comparison of the contribution of the synthetic character versus the Ground Truth (GT) character for the class $t=1$. The prediction curve denotes the classification accuracy for class $t=1$,  and consequently, the prediction curve denotes the upper bound for the attributions. }
    \label{fig:figure_synthetic}
\end{figure}
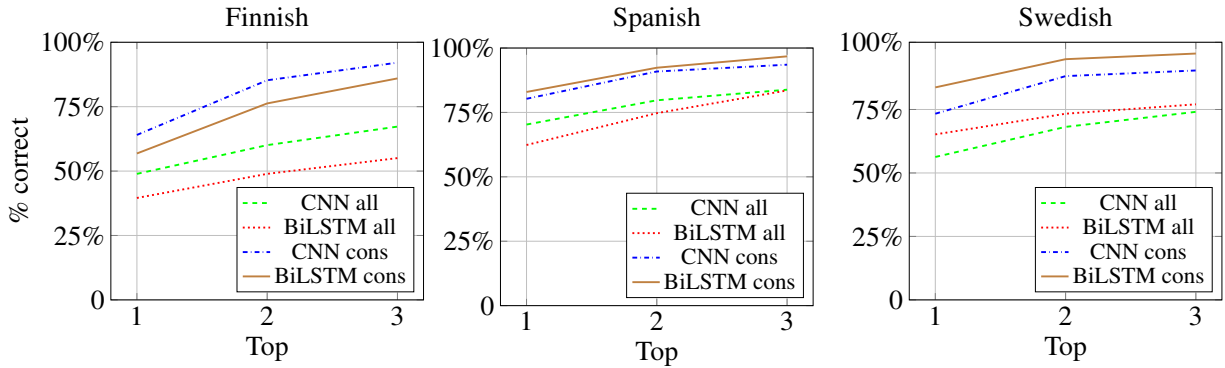
\begin{figure*}[t]
    \centering
    \begin{subfigure}[t]{0.3\textwidth}
    \centering
        \begin{tikzpicture}[scale=0.6]
\begin{axis}[
    xlabel={Top},
    ylabel={\% correct},
    title = {\LARGE Finnish},
    legend pos = south east,
    ymin = 0, ymax = 278,grid=major,label style={font=\LARGE},tick label style={font=\LARGE},
    legend style={font=\Large},xtick={0,1,2},xticklabels={1,2,3},ytick={0,69.5,139,208.5,278},yticklabels={0,{25\%},{50\%},{75\%},{100\%}} ]
\addplot [dashed, green,very thick] coordinates {(0, 136) (1, 167) (2, 187) };
\addplot [dotted, red,very thick] coordinates {(0, 110) (1, 136) (2, 153) };
\addplot [dash dot,  blue,very thick] coordinates {(0, 178) (1, 237) (2, 256) };
\addplot [solid, brown,very thick] coordinates {(0, 158) (1, 212) (2, 239) };
\legend{CNN all, BiLSTM all, CNN cons, BiLSTM cons}
\end{axis}
\end{tikzpicture}
    \end{subfigure}\hspace{0.03\textwidth}
    \centering
    \begin{subfigure}[t]{0.3\textwidth}
    \centering
        \begin{tikzpicture}[scale=0.6]
\begin{axis}[
    xlabel={Top},
    ylabel={\ },
        title = {\LARGE Spanish},
    legend pos = south east,
    ymin = 0, ymax = 340,grid=major,label style={font=\LARGE},tick label style={font=\LARGE},
    legend style={font=\Large},xtick={0,1,2},xticklabels={1,2,3},ytick={0,85,170,255,340},yticklabels={0,{25\%},{50\%},{75\%},{100\%}} ]
\addplot [dashed, green,very thick] coordinates {(0, 239) (1, 271) (2, 285) };
\addplot [dotted, red,very thick] coordinates {(0, 212) (1, 254) (2, 284) };
\addplot [dash dot,  blue,very thick] coordinates {(0, 273) (1, 309) (2, 318) };
\addplot [solid, brown,very thick] coordinates {(0, 282) (1, 314) (2, 329) };
\legend{CNN all, BiLSTM all, CNN cons, BiLSTM cons}
\end{axis}
\end{tikzpicture}
    \end{subfigure}\hspace{0.03\textwidth}    \centering
    \begin{subfigure}[t]{0.3\textwidth}
    \centering
        \begin{tikzpicture}[scale=0.6]
\begin{axis}[
    xlabel={Top},
    ylabel={\ },
        title = {\LARGE Swedish},
    legend pos = south east,
    ymin = 0, ymax = 137,grid=major,label style={font=\LARGE},tick label style={font=\LARGE},
    legend style={font=\Large},xtick={0,1,2},xticklabels={1,2,3},ytick={0,33.75,67.5,101.25,137},yticklabels={0,{25\%},{50\%},{75\%},{100\%}}]
\addplot [dashed, green,very thick] coordinates {(0, 76) (1, 92) (2, 100) };
\addplot [dotted, red,very thick] coordinates {(0, 88) (1, 99) (2, 104) };
\addplot [dash dot, blue,very thick] coordinates {(0, 99) (1, 119) (2, 122) };
\addplot [solid, brown,very thick] coordinates {(0, 113) (1, 128) (2, 131) };
\legend{CNN all, BiLSTM all, CNN cons, BiLSTM cons}
\end{axis}
\end{tikzpicture}
    \end{subfigure}
\caption{Evaluation of the attributions of CNN and BiLSTM models on the three different languages. }
\label{fig:evaluation_wordpairs}
\end{figure*}
To verify that the contextual decomposition of CNNs works correctly, we devise an experiment in which we add a synthetic token to a word of a certain class, testing whether this token gets a high attribution score with respect to that specific class.

Given a word $w$ and a corresponding binary label $t$, we add a synthetic character $c$ to the beginning of word $w$ with probability $p_{syn}$ if that word belongs to the class $t=1$ and with probability $1-p_{syn}$ if that word belongs to the class $t=0$. Consequently, if $p_{syn}=1$, the model should predict the label with a 100\% accuracy, thus attributing this to the synthetic character $c$. When $p_{syn}=0.5$, the synthetic character does not provide any additional information about the label $t$, and $c$ should thus have a small contribution.
\paragraph{Experimental setup} 
We train a CNN model on the Spanish dataset and only use words having the morphological label \emph{number}. This label has two classes \emph{plur} and \emph{sing}, and assign those classes to the binary labels zero and one, respectively. Furthermore, we add a synthetic character to each word with probability $p_{syn}$, varying $p_{syn}$ from $1$ to $0.5$ with steps of $0.1$. 
We selected 112 unique word/feature pairs from our test set with label \emph{sing} or \emph{plur}. While plurality is marked by the suffix \emph{s}, a variety of suffixes are used for the singular form. Therefore, we focus on the latter class ($t=1$). The corresponding suffix is called the Ground Truth (GT) character.

To measure the impact of $p_{syn}$, we add a synthetic character to each word of the class $t=1$ and calculate the contribution of each character by using the CD algorithm. We run the experiment five times with a different random seed and report the average correct attribution. The attribution is correct if the contribution of the synthetic/GT character is the highest contribution of all character contributions.

\paragraph{Results}
The results of our evaluation are depicted in Figure~\ref{fig:figure_synthetic}. When $p_{syn}=1$, all words of the class $t=1$ contain the synthetic character, and consequently, the accuracy for predicting $t=1$ is indeed 100\%. Moreover, the correct prediction is effectively attributed to the synthetic character (`syn. char attr.' in Figure~\ref{fig:figure_synthetic} at 100\%), with the GT character being deemed irrelevant. When the synthetic character probability  $p_{syn}$ is lowered, the synthetic character is less trustworthy and the GT character becomes more important (increasing `GT char attr.' in Figure~\ref{fig:figure_synthetic}). Finally, when $p_{syn}=0.5$, the synthetic character is equally plausible in both classes. Hence, the contribution of the synthetic character becomes irrelevant and the model attributes the prediction to other characters.

Consequently, we  can conclude that whenever there is a clear character-level pattern, the model learns the pattern and the CD algorithm is able to accurately attribute it to the correct character. 
\begin{table}[t]
\centering
\caption{Average accuracy of all models trained on Finnish, Spanish and Swedish for the task of morphological feature prediction for all unique words in the full UD test set.}
\begin{tabular}{cccc}
\toprule
& \textbf{Finnish} & \textbf{Spanish} & \textbf{Swedish} \\
\midrule
\textbf{Maj. Vote} &82.20\% &	 72.39\% & 69.79\% \\
\textbf{CNN} & 94.81\% &	88.93\% &	\textbf{90.09\%}\\
\textbf{BiLSTM} & \textbf{95.13\%} &	\textbf{89.33\%} &	89.45\%  \\
\bottomrule
\end{tabular}
\label{tab:morph_tagging_average}
\end{table}

\subsection{Evaluation of character-level attribution}
In this section, we measure and analyze (1) which characters contribute most to the final prediction of a certain label and (2) whether those contributions coincide with our linguistic knowledge about a language. To that end, we train a model to predict morphological features, given a particular word. The model does not have prior word segmentation information and thus needs to discover useful character patterns by itself. After training, we calculate the attribution scores of each character pattern within a word with respect to the correct feature class using CD, and evaluate whether this coincides with the ground truth attribution.

\paragraph{Model} We train CNN and BiLSTM models on Finnish, Spanish and Swedish. 
The average accuracies on the full test set are reported in Table~\ref{tab:morph_tagging_average}. The results for the individual classes can be found in Appendix~\ref{app:morph_features}.
As a reference for the trained models' ability to predict morphological feature classes, we  provide a naive baseline, constructed from the majority vote for each feature type.

Overall, our neural models yield substantially higher average accuracies than the baseline and perform very similar. Consequently, both the CNN and LSTM models learned useful character patterns for predicting the correct morphological feature classes. Hence, this raises the question whether these patterns coincide with our linguistic knowledge.

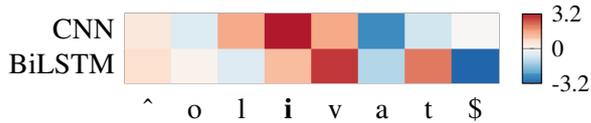
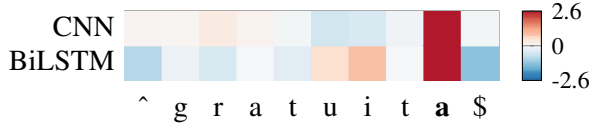
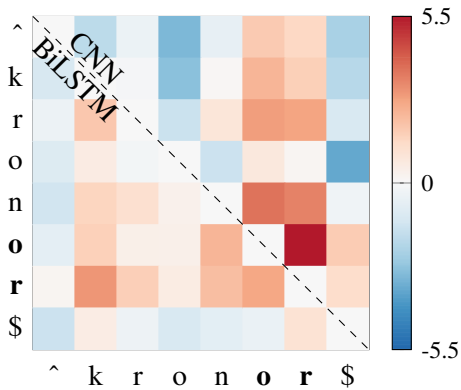
\begin{figure}[t]
	\centering
    \begin{subfigure}[b]{0.45\textwidth}
      \centering
      \begin{tikzpicture}
\begin{axis}[ymin=-0.5,ymax=1.5,xmin=-0.5,xmax=7.5,point meta min = -1,xtick={0,1,2,3,4,5,6,7,8,9,10,11},point meta max = 1,yticklabels={BiLSTM,CNN},ytick={0,1},xticklabels={\^{},o,l,\textbf{i},v,a,t,\$},height=2.5cm,width=6.5cm, 
colorbar, colorbar style={
		width=0.25cm,
        ticklabel style = {font=\small},
        ytick={-1,0,1},
        yticklabels={-3.2,0,3.2} }, 
colormap={reverse rgbu}{
        indices of colormap={
            \pgfplotscolormaplastindexof{RdBu-9},...,0 of {RdBu-9}}
    },typeset ticklabels with strut,]
\addplot [matrix plot*,point meta=explicit] file [meta=index 2] {figure_example_fi_conv.dat};
\end{axis}
\end{tikzpicture}
      \caption{Example of Finnish. Word (verb): \emph{ol\textbf{i}vat (were)}, target class: \emph{Tense=Past}}
      \label{fig:figure_example_fi}
    \end{subfigure}
    \vfill
    \begin{subfigure}[b]{0.45\textwidth}
      \centering
      \begin{tikzpicture}
\begin{axis}[ymin=-0.5,ymax=1.5,xmin=-0.5,xmax=9.5,point meta min = -1,xtick={0,1,2,3,4,5,6,7,8,9},point meta max = 1,yticklabels={BiLSTM,CNN},ytick={0,1},xticklabels={\^{},g,r,a,t,u,i,t,\textbf{a},\$},height=2.5cm,width=6.5cm,  
colorbar, colorbar style={
		width=0.25cm,
        ticklabel style = {font=\small},
        ytick={-1,0,1},
        yticklabels={-2.6,0,2.6} }, 
colormap={reverse rgbu}{
        indices of colormap={
            \pgfplotscolormaplastindexof{RdBu-9},...,0 of {RdBu-9}}
    },typeset ticklabels with strut,]
\addplot [matrix plot*,point meta=explicit] file [meta=index 2] {figure_example_es.dat};
\end{axis}
\end{tikzpicture}
      \caption{Example of Spanish. Word (adjective): \emph{gratuit\textbf{a} (free)}, target: \emph{Gender=Fem}.}
      \label{fig:figure_example_es}
  \end{subfigure}
  \begin{subfigure}[b]{0.45\textwidth}
  	  \centering
      \begin{tikzpicture}
\begin{axis}[ymin=-0.5,ymax=7.5,xmin=-0.5,xmax=7.5,point meta min = -1,xtick={0,1,2,3,4,5,6,7},point meta max = 1,yticklabels={\$,\textbf{r},\textbf{o},n,o,r,k,\^{}},ytick={0,1,2,3,4,5,6,7},xticklabels={\^{},k,r,o,n,\textbf{o},\textbf{r},\$},height=6cm,width=6cm,  
colorbar, colorbar style={
		width=0.25cm,
        ticklabel style = {font=\small},
        ytick={-1,0,1},
        yticklabels={-5.5,0,5.5} }, 
colormap={reverse rgbu}{
        indices of colormap={
            \pgfplotscolormaplastindexof{RdBu-9},...,0 of {RdBu-9}}
    },typeset ticklabels with strut,]
\addplot [matrix plot*,point meta=explicit] file [meta=index 2] {figure_example_sv_bigram.dat};
\draw [dashed] (rel axis cs:0,1) -- (rel axis cs:1,0);
\node[rotate=-45] at (1,6.5) {CNN};
\node[rotate=-45] at (0.5,6) {BiLSTM};
\end{axis}
\end{tikzpicture}
      \caption{Example of Swedish. Word (noun): \emph{kron\textbf{or} (Swedish valuta as in dollars)}, target: \emph{Number=Plur}. }
      \label{fig:figure_example_sv}
   \end{subfigure}
         \caption{Character-level contributions for predicting a particular class. Positive contributions are highlighted in red and negative contributions in blue. The ground truth character sequence is highlighted in bold.}
         \label{fig:viz_attributions}
\end{figure}

\paragraph{Evaluation}
For each annotated word/feature pair, 
we measure if the ground truth character sequence corresponds to the set or sequence of characters with the same length within the considered word that has the highest contribution for predicting the correct label for that word.

In the first setup, we only compare with character sequences having a consecutive set of characters (denoted \emph{cons}). In the second setup, we compare with any set of characters (denoted \emph{all}). We rank the contributions of each character set and report top one, two, and three scores. Because start-of-word and end-of-word characters are not annotated in the dataset, we do not consider them part of the candidate character sets.

\paragraph{Results}
The aggregated results for all classes and character sequence lengths are shown in Figure~\ref{fig:evaluation_wordpairs}. In general, we observe that for almost all models and setups, the contextual decomposition attribution coincides with the manually-defined segmentations for at least half of the word/feature pairs. When we only consider the top two consecutive sequences (marked as \emph{cons}), accuracies range from 76\% up to 93\% for all three languages. For Spanish and Swedish, the top two accuracies for character sets (marked as \emph{all}) are still above 67\%, despite the large space of possible character sets, whereas all ground truth patterns are consecutive sequences. While the accuracy for Finnish is lower, the top two accuracy is still above 50\%.

Examples for Finnish, Spanish and Swedish are shown in Figure~\ref{fig:viz_attributions}. For Finnish, the character with the highest contribution \emph{i} coincides with the ground truth character for the CNN model. This is not the case for the BiLSTM model which focuses on the character \emph{v}, even though the correct label is predicted. For Spanish, both models strongly focus on the ground truth character \emph{a} for predicting the feminine gender. For Swedish, the ground truth character sequence is the suffix \emph{or} which denotes plurality. Given that \emph{or} consists of two characters, all contributions of character sets of two characters are visualized.  
As can be seen, the most important set of two characters is \{o,r\} for the CNN and \{k,r\} for the BiLSTM model. However, \{o,r\} is the second most important character set for the BiLSTM model. Consequently, the BiLSTM model deemed the interaction between a root and suffix character more important than between two suffix characters.

\begin{table*}[htb]
\centering
\caption{The most frequent character sets used by a model for predicting a specific class. The frequency of occurrence is shown between brackets. An underscore denotes an unknown character. }
\begin{tabular}{ C{1cm}cC{3cm}C{3.5cm}C{3.7cm}C{1.8cm}}
\toprule
&& \textbf{One character} & \textbf{Two characters} & \textbf{Three characters} & \textbf{Examples}\\
\midrule
\multirow{-4}{*}{\rotatebox[origin=c]{90}{\parbox{4cm}{\center{\textbf{Finnish}\\Tense=Past} }}}
& BiL. & i (69\%), t (22\%), v (4\%), a (2\%) & ti (13\%), t\_i (12\%), v\_t (9\%), ui (6\%) & tti (8\%), iv\_t (5\%), t\_\_ti (3\%), sti (3\%)& ol\textbf{iv}a\textbf{t},   n\"{a}y\textbf{tti}k\"{a}\"{a}n \\
\cline{2-6}
& CNN & i (71\%), t (8\%), s (6\%), o (5\%) & ui (12\%), si (11\%), ti (11\%), oi (9\%) &a\_\_ui (3\%), tii (3\%), iv\_\_\$ (2\%), ui\_\_t (2\%)&  tie\textbf{si}, me\textbf{i}d\"{a}t \\
\midrule
\multirow{-4}{*}{\rotatebox[origin=c]{90}{\parbox{4cm}{\center{\textbf{Spanish}\\Gend=Fem}} }}
& BiL. & a (69\%), i (16\%), d (6\%), e (4\%) & as (23\%), a\$ (13\%), ad (7\%), ia (5\%) & ia\$ (4\%), ad\$ (3\%), da\$ (3\%), ca\$ (2\%) & toleranc\textbf{ia}, ciud\textbf{ad} \\ 
\cline{2-6}
& CNN & a (77\%), \'{o} (14\%), n (4\%), d (3\%) & a\$ (34\%), as (20\%), da (8\%), i\'{o}  (7\%) &dad (5\%), da\$ (4\%), a\_i\'{o}  (4\%), si\'{o}  (2\%) &  firm\textbf{as}, preci\textbf{si\'{o}}n \\
\midrule
\multirow{-4}{*}{\rotatebox[origin=c]{90}{\parbox{4cm}{\center{\textbf{Swedish}\\Numb=Plur}} }}
& BiL. & n (25\%), r (19\%), a (14\%), g (7\%) &na (13\%), a\_\_r (4\%), or (3\%), n\_\_r (3\%) & iga (5\%), rna (3\%), ner (1\%), der (1\%) & kron\textbf{or}, perio\textbf{der} \\
\cline{2-6}
& CNN &n (21\%), a (18\%), r (15\%), d (5\%) & rn (8\%), na (5\%), or (4\%), er (3\%) & rna (7\%), arn (3\%), iga (2\%), n\_ar (2\%) & krafte\textbf{rna}, sak\textbf{er} \\
\bottomrule

\end{tabular}
\label{tab:example_grams}
\end{table*}
\subsection{Analysis of learned patterns}
In the previous section, we showed that there is a strong relationship between the manually-defined morphological segmentation and the patterns a neural network learns. However, there is still an accuracy gap between the results obtained using consecutive sequences only and results obtained using all possible character sets. Hence, this leads to the question which patterns the neural network focuses on, other than the manually defined patterns we evaluated before.
To that end, for each of the three languages, we selected a morphological class of interest and evaluated for all words in the full UD test set that were assigned to that class what the most important character set of length one, two and three was. In other words, we evaluated for each word for which the class was correctly predicted, which character set had the highest positive contribution towards predicting that class. The results can be found in Table~\ref{tab:example_grams}.

\paragraph{Finnish} In Finnish, adding the suffix \emph{i} to a verb, transforms it in the past tense. Sometimes the character \emph{s} is added, resulting in the suffix \emph{si}. The latter is a frequently used bigram pattern by the CNN but less by the BiLSTM. The BiLSTM combines the suffix \emph{i} with another suffix \emph{vat} which denotes third person plural in the character pattern \emph{iv\_t}. 

\paragraph{Spanish} While there is no single clear-cut rule for the Spanish gender, in general the suffix \emph{a} denotes the feminine gender in adjectives. However, there exist many nouns that are feminine but do not have the suffix \emph{a}. \citet{teschner} identify \emph{d}, and \emph{i\emph{\'{o}}n} as typical endings of feminine nouns, which our models identified too as for example \emph{ad\$} or \emph{i\'{o}}/\emph{si\'{o}}.

\paragraph{Swedish} In Swedish, there exist four suffixes for creating a plural form: \emph{or}, \emph{ar}, \emph{(e)r} and \emph{n}. Both models identified the suffix \emph{or}. However, similar to Finnish, multiple suffixes are merged. In Swedish, the suffix \emph{na} only occurs together with one of the first three plural suffixes. Hence, both models correctly identified this pattern as an important pattern for predicting the class number=plural, rather than the linguistically-defined pattern.

\subsection{Interactions of learned patterns}
In the previous section, the pattern \emph{a\$} showed to be the most important pattern in 34\% of the correctly-predicted feminine Spanish words in our dataset. However, there exist many words that end with the character \emph{a} that are not feminine.  For example the third person singular form of the verb \emph{gustar} is \emph{gusta}. Hence, this raises the question if the model will classify \emph{gusta} wrongly as feminine or correctly as NA. As an illustration of the applicability of CD for morphological analysis, we will study this case in more detail.

From the full UD test set, we selected all words that end with the character \emph{a} and that do not belong to the class gender=feminine. 
Using the Spanish CNN model, we predicted the gender class for each word and divided the words into two groups: predicted as feminine and predicted as not-feminine (\_NA\_ or masculine). The resulted in 44 and 199 words. Next, for each word in both groups we calculated the most positively and negatively contributing character set out of all possible character sets of any length within the considered word, using the CD algorithm. We compared the contribution scores in both groups using a Kruskal-Wallis significance test.\footnote{The full statistical analysis is provided in Appendix~\ref{app:statistical}.}
While no significant (p < 0.05) difference could be found between the positive contributions of both groups (p=1.000), a borderline significant difference could be found between the negative contributions of words predicted as feminine and words predicted as not-feminine (p=0.070). 

Consequently, the CNN model's classification decision is based on finding enough negative evidence to counteract the positive evidence found in the pattern \emph{a\$}, which CD was able to uncover. 

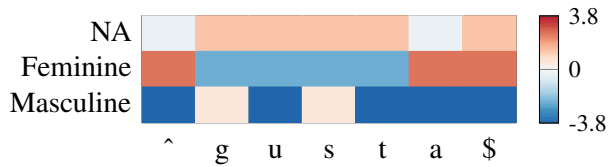
\begin{figure}[t]
\centering
\begin{tikzpicture}
\begin{axis}[ymin=-0.5,ymax=2.5,xmin=-0.5,xmax=6.5,point meta min = -1,xtick={0,1,2,3,4,5,6,7,8,9,10,11},point meta max = 1,yticklabels={Masculine,Feminine,NA},ytick={0,1,2},xticklabels={\^{},g,u,s,t,a,\$},height=3cm,width=6.5cm, 
colorbar, colorbar style={
		width=0.25cm,
        ticklabel style = {font=\small},
        ytick={-1,0,1},
        yticklabels={-3.8,0,3.8} }, 
colormap={reverse rgbu}{
        indices of colormap={
            \pgfplotscolormaplastindexof{RdBu-9},...,0 of {RdBu-9}}
    },typeset ticklabels with strut,]
\addplot [matrix plot*,point meta=explicit] file [meta=index 2] {figure_negation_example.dat};
\end{axis}
\end{tikzpicture}
\caption{Visualization of the most positively and negatively contributing character set for each class of the morphological feature class gender for the Spanish verb \emph{gusta} (likes).}
\label{fig:negation}
\end{figure}

A visualization of this interaction is shown in Figure~\ref{fig:negation} for the word \emph{gusta}. While the positive evidence is the strongest for the class feminine, the model identifies the verb stem \emph{gust} as negative evidence which ultimately leads to the correct final prediction NA.

\section{Conclusion}
While neural network-based models are part of many NLP systems, little is understood on how they handle the input data. 
We investigated how specific character sequences at the input of a neural network model contribute to word-level tagging decisions at the output, and if those contributions follow linguistically interpretable rules.

First, we presented an analysis and visualization technique to decompose the output of CNN models into separate input contributions, based on the principles outlined by \citet{james2018beyond} for LSTMs.
This allowed us then to quantitatively and qualitatively compare the character-level patterns the CNNs and BiLSTMs learned for the task of morphological tagging. We showed that these patterns generally coincide with the morphological segments as defined by linguists for three morphologically different languages, but that sometimes other linguistically plausible patterns are learned. Finally, we showed that our CD algorithm for CNNs is able to explain why the model made a wrong or correct prediction.

By visualizing the contributions of each input unit or combinations thereof, we believe that much can be learned on how a neural network handles the input data, why it makes certain decisions, or even for debugging neural network models.

\section*{Acknowledgments}
The authors would like to thank the anonymous reviewers and members of IDLab for their valuable feedback. FG would like to thank Kim Bettens for helping out with the statistical analysis.

The research activities as described in this paper were funded by Ghent University, imec, Flanders Innovation \& Entrepreneurship (VLAIO), the Fund for Scientific Research-Flanders (FWO-Flanders), and the European Union.

\bibliographystyle{acl_natbib_nourl}

\clearpage
\section*{Appendix}
\appendix

\section{Notes on Dataset Selection}\label{app:dataset}
In the paper \cite{silfverberg} that introduced the morphological segmentations for a subset of the Universal Dependencies dataset 1.4, sentences from the training dataset were selected for constructing the test set. Although the paper mentions that all test set sentences for Finnish, Spanish and Swedish were selected from the Universal Dependencies test sets, this is not the case for Finnish. The first 515 lines of \emph{fi-ud-train.conllu} were used for selecting 300 test set words. Given that a new sentence starts at line 521, we removed the first 520 lines from the Finnish training set. This is only 0.3\% of the full training set, and consequently, this will have a negligible impact on our conclusions. Note that, for Spanish and Swedish, the segmented words were indeed selected from their respective test sets. The above observations were also confirmed by the first author of the original paper.

\section{Overview Morphological Classes Used}\label{app:morph_classes_used}
In Table \ref{tab:classes_fi}, Table \ref{tab:classes_es} and Table \ref{tab:classes_sv}, all class types and corresponding feature class values for Finnish, Spanish and Swedish are listed. During training, each class type has a specific multinomial regression layer which predicts a single value for that class type. However, all class types are jointly trained.

\begin{table}[tp]
\centering
\caption{Overview of classes used for Finnish.}
\begin{tabular}{ l p{5cm}}
\toprule
Class type & Values  \\ 
\midrule
Number & \_NA\_    Sing    Plur\\
PartForm    &    \_NA\_   Past    Pres    Agt     Neg\\
Case &   \_NA\_    Ela     Ine     Ins     Par     Ill     Com     Nom     All     Acc    Ade     Gen     Ess     Abl   Tra Abe \\
Person  & \_NA\_    1       2       3\\
Derivation  &    \_NA\_    Ja      Minen   Sti     Vs      Tar     Llinen  Inen    U      Ttaa    Ttain   Lainen  Ton\\
Person[psor] &   \_NA\_   1       2       3\\
VerbForm  &      \_NA\_    Inf     Part    Fin\\
Mood   & \_NA\_   Imp     Cnd     Pot     Ind\\
Tense  & \_NA\_    Past    Pres\\
Clitic & \_NA\_    Pa,S    Han     Ko      Pa      Han,Pa  Han,Ko  Ko,S    S       Kin     Kaan    Ka\\
Degree & \_NA\_  Pos     Cmp     Sup\\
Voice  &\_NA\_    Pass    Act\\

\bottomrule
\end{tabular}
\label{tab:classes_fi}
\end{table}

\begin{table}[tp]
\centering
\caption{Overview of classes used for Spanish.}
\begin{tabular}{ l l}
\toprule
Class type & Values  \\ 
\midrule
Person  & \_NA\_           1       2	3\\
Mood   &\_NA\_    Imp     Ind     Sub     Cnd\\
Tense   &\_NA\_ Fut     Imp     Pres    Past\\
Gender  &\_NA\_    Fem     Masc\\
VerbForm &       \_NA\_    Inf     Ger     Part    Fin\\
Number & \_NA\_   Sing Plur\\
\bottomrule
\end{tabular}
\label{tab:classes_es}
\end{table}

\begin{table}[tp]
\centering
\caption{Overview of classes used for Swedish.}
\begin{tabular}{ l l}
\toprule
Class type & Values  \\ 
\midrule
Gender&  \_NA\_    Neut    Masc    Fem     Com\\ 
Degree&  \_NA\_    Sup     Cmp     Pos\\ 
Number&  \_NA\_    Sing Plur\\ 
Case  &  \_NA\_    Gen     Nom     Acc\\ 
Poss &   \_NA\_    Yes\\ 
Voice &  \_NA\_    Act     Pass\\ 
Tense & \_NA\_    Pres    Past\\ 
Definite  &      \_NA\_    Ind     Def\\ 
VerbForm   &     \_NA\_    Sup     Part    Inf     Fin     Stem\\ 
\bottomrule
\end{tabular}
\label{tab:classes_sv}
\end{table}

\section{Individual Results Morphological Tagging}\label{app:morph_features}
In Table~\ref{tab:morph_tagging_fi}, Table~\ref{tab:morph_tagging_es} and Table~\ref{tab:morph_tagging_sv}, the individual results for each morphological feature class for Finnish, Spanish and Swedish can be found.

\section{Full statistical analysis for "Interactions of learned patterns"}\label{app:statistical}
From the full UD test set, we selected all words that end with the character \emph{a} and evaluated the morphological feature type gender for all of them. We selected three groups: 
\begin{itemize}
\item Words that have the label gender=feminine and are classified as gender=feminine, called wf\_pf. This group contains 219 words.
\item Words that do not have the label gender=feminine are classified as gender=feminine,  wnf\_pf. This group contains 44 words.
\item Words that do not have the label gender=feminine are classified as either gender=NA or gender=masc, i.e. not-feminine, called wnf\_pnf. This group contains 199 words.
\end{itemize}

For each group, we calculated the contributions of all possible character sets of different length within each word and selected the highest contribution score and the lowest contribution score for each word. In other words, we look for the sets of characters that generate the strongest positive and negative contributions for predicting the class gender=feminine. These two contribution scores are the determining factors for certain classification decisions.
\subsection{Maximum contribution scores}
Based on a Kruskal-Wallis test, a statistically significant difference was found between the three groups, $H(2)=50,600$, $p<0.001$. Pairwise comparisons with adjusted p-values showed no significant difference in positive contributions scores between the groups wnf\_pf and wnf\_pnf ($p=1.000$). Hence, non-feminine words have similar positive contribution scores, independent of the classification result. Furthermore, significant differences were found between the positive contribution scores of the groups wf\_pf and wnf\_pf ($p < 0.001$) and the groups wf\_pf and wnf\_pnf ($p < 0.001$), indicating a difference between the positive contributions of feminine words and non-feminine words.

\subsection{Minimum contribution scores}
Based on a Kruskal-Wallis test, an overall statistically significant difference was found between the three groups, $H(2) = 36.710$, $p < 0.001$. Pairwise comparisons with adjusted p-values showed that there was no significant difference between the groups wf\_pf and  wnf\_pf ($p = 0.585$), showing that the negative contribution scores of words classified as feminine are similar despite that the fact that the words from wnf\_pf are not feminine. A strong significant difference was found between the groups wf\_pf and wnf\_pnf ($p < 0.001$) and a borderline significant difference between the groups wnf\_pnf and wnf\_pf ($p < 0.070$). Consequently, there is a clear difference between the negative contributions of non-feminine words that are classified as not-feminine and words that are classified as feminine. Moreover, words that are wrongly classified as feminine have similar negative contribution scores as words classified correctly as feminine.

\begin{table*}[htbp]
\centering
\caption{Per class accuracy on the Finnish test set.}
\begin{tabular}{ccccccc}
\toprule
& Number & Partform & Case & Person & Derivation & Person[psor] \\
\midrule
Maj. Vote & 64.42\% & 94.33\% & 28.49\% & 89.17\% & 98.43\% & 96.05\% \\
CNN & 89.40\% & 96.97\% & 87.00\% & 95.81\% & 99.07\% & 98.49\% \\ 
BiLSTM & 89.67\% & 97.86\% & 87.89\% & 95.77\% & 99.11\% & 99.29\% \\

\bottomrule
\end{tabular}

\begin{tabular}{ccccccc}
\toprule
 & Verbform & Mood & Tense & Clitic & Degree & Voice \\
\midrule
Maj. Vote & 77.54\% & 87.77\% & 89.09\% & 98.49\% & 84.16\% & 78.49\% \\
CNN & 93.05\% & 95.90\% & 96.17\% & 99.51\% & 92.70\% & 93.59\% \\ 
BiLSTM & 93.19\% & 96.13\% & 95.99\% & 99.51\% & 92.97\% & 94.12\% \\
\bottomrule
\end{tabular}

\label{tab:morph_tagging_fi}
\end{table*}

\begin{table*}[htbp]
\centering
\caption{Per class accuracy on the Spanish test set.}
\begin{tabular}{ccccccc}
\toprule
& Person & Mood & Tense & Gender & Verbform & Number \\
\midrule
Maj. Vote & 85.26\% & 87.62\% & 85.99\% & 54.40\% & 75.49\% & 45.56\% \\
CNN &91.84\% & 93.51\% & 91.11\% & 84.62\% & 88.08\% & 84.41\% \\
BiLSTM & 91.95\% & 93.41\% & 90.90\% & 84.31\% & 89.02\% & 86.40\% \\ 

\bottomrule
\end{tabular}

\label{tab:morph_tagging_es}
\end{table*}

\begin{table*}[htbp]
\centering
\caption{Per class accuracy on the Swedish test set.}
\begin{tabular}{cccccccccc}
\toprule
& Gender & Degree & Number & Case & Poss & Voice & Tense & Definite & Verbform \\
\midrule
Maj. Vote & 46.64\% & 84.57\% & 42.21\% & 62.73\% & 99.67\% & 83.99\% & 87.57\% & 41.54\% & 79.19\% \\
CNN & 86.18\% & 93.78\% & 79.45\% & 87.79\% & 99.94\% & 94.60\% & 94.29\% & 83.83\% & 90.98\%  \\ 
BiLSTM & 83.97\% & 94.26\% & 78.72\% & 86.04\% & 99.97\% & 93.75\% & 93.84\% & 83.86\% & 90.64\% \\

\bottomrule
\end{tabular}

\label{tab:morph_tagging_sv}
\end{table*}

\end{document}